\documentclass{article}

\usepackage{PRIMEarxiv}

\usepackage[utf8]{inputenc} 
\usepackage[T1]{fontenc}    
\usepackage{hyperref}       
\usepackage{url}            
\usepackage{booktabs}       
\usepackage{amsfonts}       
\usepackage{nicefrac}       
\usepackage{microtype}      
\usepackage{lipsum}
\usepackage{fancyhdr}       
\usepackage{graphicx}       
\graphicspath{{media/}}     

\title{Online anomaly detection based on \\ Reservoir Sampling and LOF for IoT devices
}

\author{
  Tomasz Szydlo \\
  Institute of Computer Science \\
  AGH University of Science and Technology \\
  Krakow, Poland\\
  \texttt{tszydlo@agh.edu.pl} \\
}

\begin{document}
\maketitle

\begin{abstract}
The growing number of IoT devices and their use to monitor the operation of machines and equipment increases interest in anomaly detection algorithms running on devices. However, the difficulty is the limitations of the available computational and memory resources on the devices. In the case of microcontrollers (MCUs), these are single megabytes of program and several hundred kilobytes of working memory. Consequently, algorithms must be appropriately matched to the capabilities of the devices. In the paper, we analyse the processing pipeline for anomaly detection and implementation of the Local Outliner Factor (LOF) algorithm on a MCU. We also show that it is possible to train such an algorithm directly on the device, which gives great potential to use the solution in real devices.
\end{abstract}


\section{Introduction}
Internet of Things (IoT) systems and devices have several applications in different aspects of our lives, including smart homes and cities, autonomous drones and industrial machines. Such applications are possible thanks to the enabling technologies, such as the miniaturisation of electronic systems, communication protocols and power sources. The growing number of IoT devices and their use to monitor the operation of industrial machines and equipment increases interest in anomaly detection algorithms running directly on devices. The difficulty is the limitations of the available computational and memory resources on the devices. In the case of microcontrollers (MCUs), these are single megabytes of program memory and several hundred kilobytes of working memory. 

As a result, the algorithms must be adapted to the computational and memory capabilities of the devices. Moreover, the problem of training these algorithms directly on devices is particularly important  due to the need to adjust the anomaly detection algorithms to the specific deployment. Thus, there are two research problems to be solved. The first relates to what class of anomaly detection algorithms can be implemented on an MCU. The second relates to training these algorithms and collecting the profile of the correct operation of the monitored device.

In the paper, we analyse the processing pipeline for anomaly detection and implementation of the Local Outliner Factor algorithm on a MCU. We also show that it is possible to train such an algorithm directly on the device, which gives great potential to use the solution in real devices.

\section{Related work}
\subsection{Anomaly Detection}
Anomaly detection is understood as the identification of events that deviate significantly from the normal state. The application of anomaly detection covers several areas, including industrial machine monitoring, financial transaction fraud detection or devices malfunction. 

Anomaly detection algorithms fall into two main classes of solutions \cite{AnomalyTaxonomy}. The first includes supervised algorithms, which require labelled data sets distinguishing between correct and incorrect states. These solutions can identify a specific problem but are challenging to implement due to the lack of sufficiently large datasets. The second group of solutions includes unsupervised algorithms, which assume that the data are not labelled. This group of algorithms is the most commonly used due to their relative ease of implementation and adaptability to a specific application.

Unsupervised anomaly detection algorithms are divided into groups depending on how abnormal states are identified. These include distance-based (k-NN, LOF, K-means), statistical (GMM, histogram-based) and classification-based (one-class classifier, isolation forest) methods. These algorithms perform well in IoT systems for monitoring industrial machines and equipment. They can be trained on data from properly functioning machines and detect possible deviations from the norm. However, a research problem is the implementation of these algorithms on devices with scarce computational and memory resources.

\subsection{TinyML}
TinyML concept is focused on running ML models on resource-constrained devices such as single board computers (e.g. Raspberry Pi or Nvidia Jetson Nano) or microcontrollers (ARM Cortex M, RISC-V) \cite{6m}. Due to the resource constraints, there are several techniques to reduce the size of neural networks. The most common are network weight quantisation, which is the conversion of floating-point numbers to fixed-point numbers, resulting in a final 25\% of original network size \cite{quantization}. Binarisation of weights \cite{binary} is an extension to this method, which further reduces memory requirements during inference. Another technique is pruning \cite{prunning}, which removes the least important weights in the neural network. Then inference using NNs on microcontrollers is possible with frameworks such as TensorFlow Lite and PyTorch Mobile. Chip vendors such as ARM, provides the libraries for efficient matrix calculations (e.g. CMSIS-NN). Another approach includes the use of NN compilers such as Apache TVM, which provides a C code generator and machine code optimiser that improves network layer operations for specific compute cores.

However, a much more difficult research problem is to apply these concepts to the  IoT devices equipped with microcontrollers \cite{tinyml-surv}. In this case, TinyML solutions involve the implementation of completely different NN architectures, preprocessing of digital signals (e.g. FFT, RMS) and the use of classical ML methods such as decision forests \cite{SzydloH10}. One of the techniques used in that case is source code generation of ML models for embedded devices (Q-Learning, Decision forrest, MLP) using tools such as FogML.

FogML\footnote{https://github.com /tszydlo/FogML} is a tool for synthesising the source code of ML models for the C language enabling inference directly on the IoT device. Using FogML tools is a process consisting of several steps. In the first step of the process, the ML model is learned using the high-level scikit-learn package. Then, based on the learned model, a dedicated generator creates C code. Finally, it becomes an integral part of the device's firmware. The FogML toolkit supports classification and reinforcement learning algorithms. In the paper, we present its extension to online anomaly detection algorithms. Moreover, the classification and anomaly detection algorithms can complement each other and be used simultaneously on devices.

\section{Concept}

\begin{figure}
  \centering
   \includegraphics[width=0.9\textwidth]{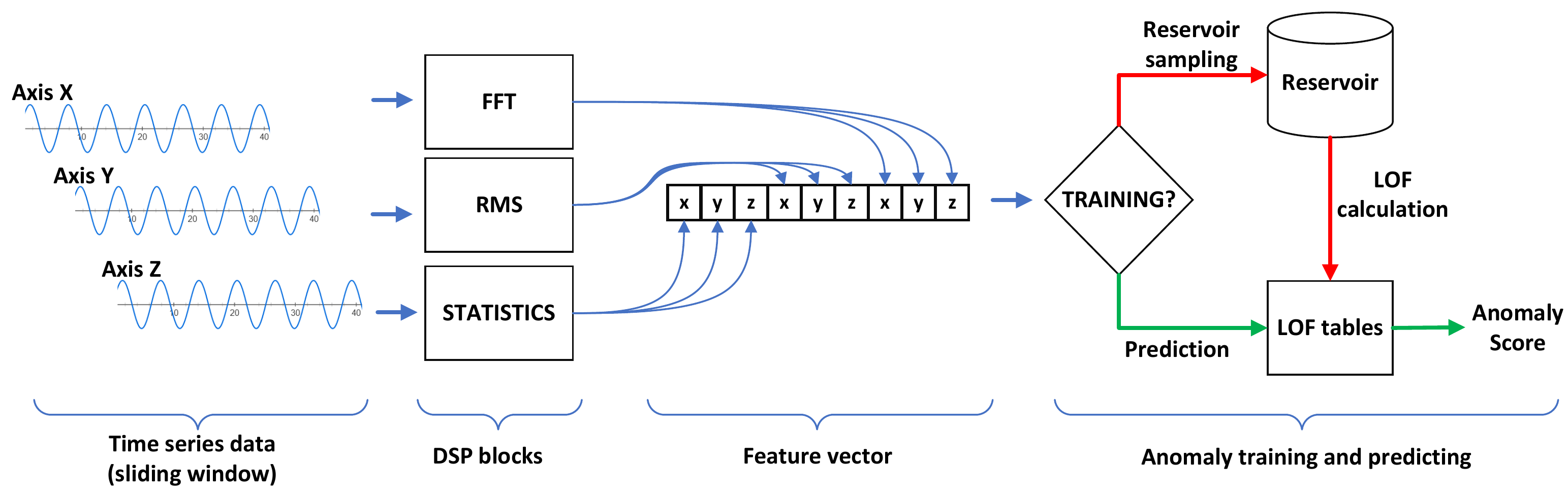}
  \caption{Processing pipeline.}
  \label{fig:pipeline}
\end{figure}

Implementing anomaly detection algorithms based on distance metrics like k-NN or LOF requires a specific approach for MCU, especially for online learning on an IoT device. In these algorithms, it is necessary to store in memory a set of points to which a new input data is compared in order to determine the value of the similarity/anomaly score. 

Fig.~\ref{fig:pipeline} depicts the concept of sensor data processing pipeline for anomaly detection. Input data in most cases are time series e.g. CPU utilization and memory consumption or accelerometer readings. For more resource rich devices such as RPi4, we would use LSTM neural networks for processing these data, but their implementation is not possible in the case of MCU. Therefore, we propose a digital signal processing~(DSP) module, which for the time series provided on input, will process it in sliding windows, and will calculate a vector containing indicators describing the input signal such as \textit{min}, \textit{max}, \textit{standard deviation}, \textit{RMS}, and \textit{FFT peaks}. 

Due to the limited memory of MCU, we propose to use the reservoir sampling method \cite{ressamp} to limit the size of the set of points. This method allows for choosing a random sample of \textit{k} items from a population of unknown size \textit{n} in a single pass over the items. We will use this method to select a \textit{k} points during the learning phase from the data stream (e.g. cpu utilization, temperature) that will be its statistical sample and, at the same time, will adjust to the length of the observed data stream. In addition, the number of \textit{k} will be set to not exceed the amount of memory available in the MCU. Then, having such \textit{k} points for a properly working system, we will calculate, e.g. the LOF or k-NN model, which we will use for assessing anomaly scores in runtime by processing system parameters and metrics.

We chose to use the LOF \cite{LOF} algorithm in the solution because of its self-adaptation to the data and returning a normalised anomaly score. LOF algorithm is an unsupervised anomaly detection method that computes local density deviation of the data point and compares it to the neighbouring ones. Lower density of the points means that these points might be considered as outliners or anomalies. The anomaly score values greater than 1.0 usually indicate the anomaly.

The use of the LOF algorithm in the presented pipeline is as follows: 
\begin{itemize}
    \item training phase - data points in the reservoir are passed to the LOF algorithm that computes point density arrays. The training process may be initiated at the start of the device operation or cyclically by updating the reservoir contents. 
    \item anomaly detection phase - new measurement data do not update the reservoir content nor density arrays but are analyzed as the so-called \textit{novelty detection}. The density around the new point is calculated assuming that the contents of the reservoir are its neighbourhood. Based on that, anomaly score is calculated.
\end{itemize}

Switching between the training and anomaly detection phases depends on the use of the device and its operating strategy.

\section{Evaluation}
\begin{figure}
  \centering
   \includegraphics[width=0.5\textwidth]{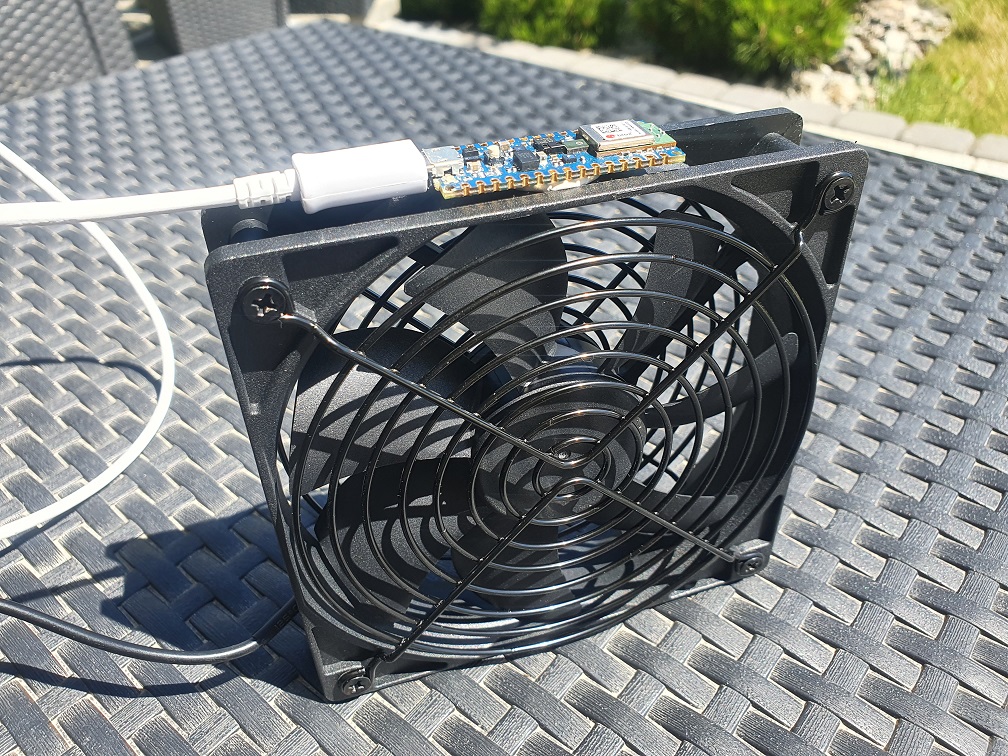}
  \caption{USB fan with Arduino 33 BLE sense attached.}
  \label{fig:fan}
\end{figure}

We have evaluated the solution on an ARM microcontroller with a Cortex M4 core, on which we processed the time series from an accelerometer installed on the computer fan. The testbed is shown in Fig.~\ref{fig:fan}. In the experiments, we have used Arduino BLE Sense 33 Nano development board equipped with the nRF52 chip and several sensors - accelerometer, magnetometer, microphone, light and gesture sensor. We conducted two types of experiments - quantitative ones to analyze the computational complexity of the processing pipeline, and qualitative ones showing how the solution works on real data.

\subsection{Quantitative tests}
\begin{figure}
  \centering
   \includegraphics[width=0.6\textwidth]{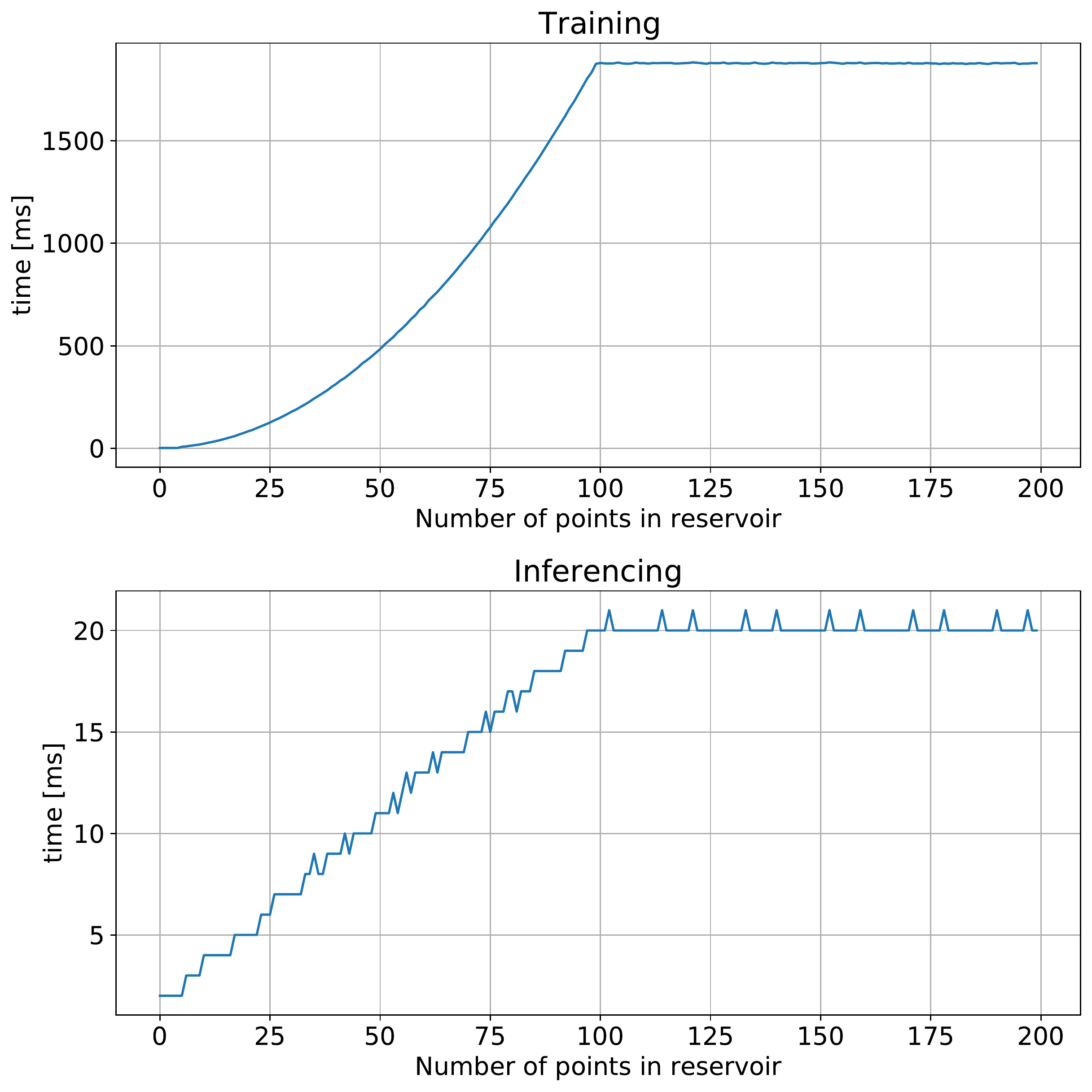}
  \caption{Processing time of the algorithm for training and inferencing phases.}
  \label{fig:complexity}
\end{figure}

The quantitative tests were conducted in a synthetic environment. We assumed that at the beginning of the experiment the reservoir was empty and its capacity was set to 100 elements. We then iteratively added more measurement points to the reservoir and performed two operations - (1) training, and (2) detecting anomalies. The results are shown in Fig.~\ref{fig:complexity}. The computational complexity of the training algorithm is of order $O(n^{2})$, while that of the inferencing is $O(n)$. For 100 points in the reservoir, the training time is about 1.8s, while the inference time is 20ms for the used MCU.

\subsection{Qualitative tests}
\begin{figure}
  \centering
   \includegraphics[width=0.7\textwidth]{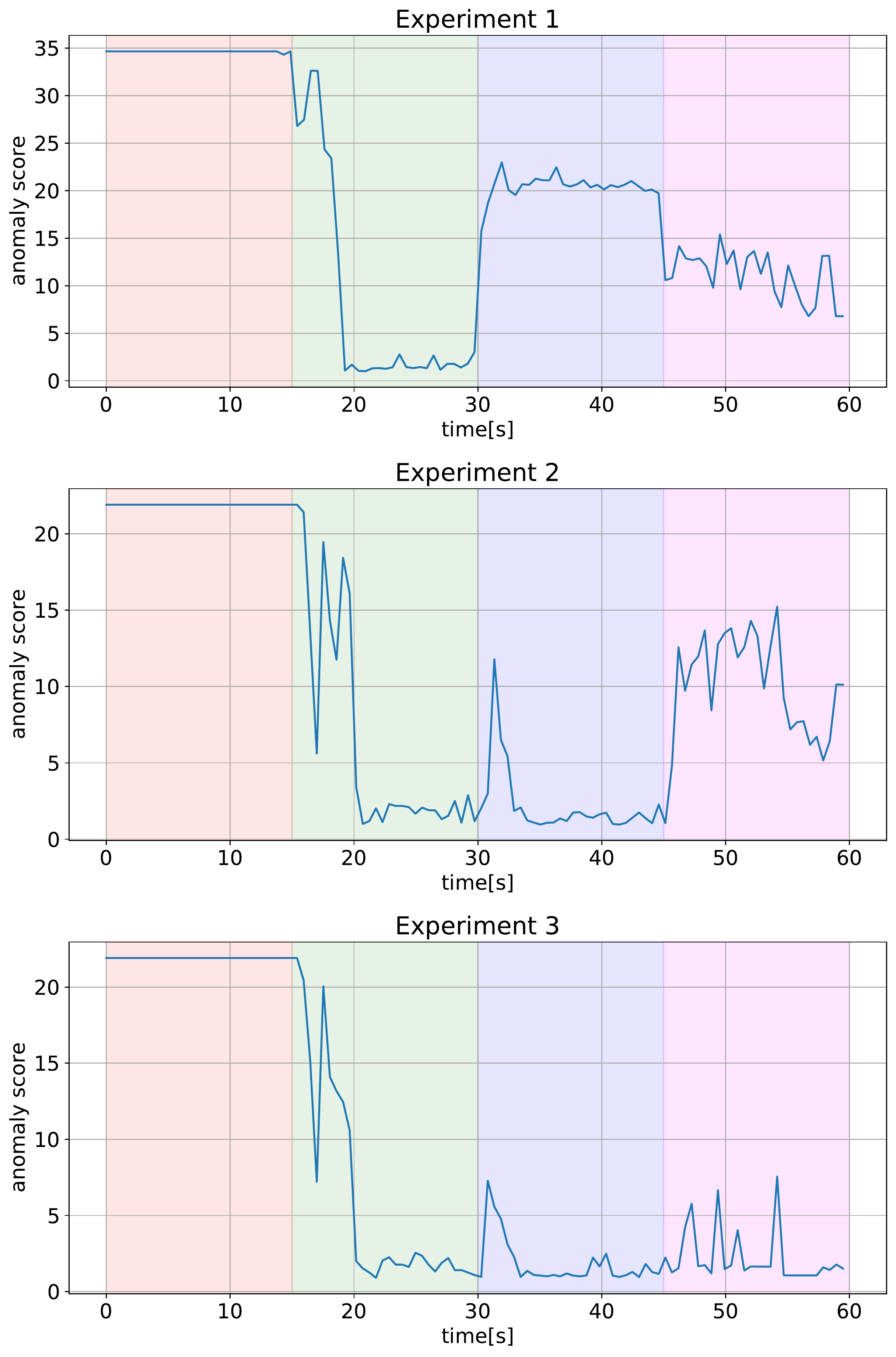}
  \caption{The value of the anomaly score during the experiment. The colour indicates the speed of the fan - red for stopped, green for speed one, purple for speed two and magenta for speed three.}
  \label{fig:exp}
\end{figure}

Qualitative experiments were performed to validate the processing pipeline applied to the real device. The fan used in the experiment had three speeds and an off state. Each of the graphs shown in Fig.~\ref{fig:exp} depicts the results of the separate experiments. Each one lasted the 60s, during which the fan speed was increased every 15s, beginning from off state.

Before the first experiment, 16 points were collected in the reservoir for speed one. During the first 15s, when the fan was off, the anomaly score value was 35, indicating an anomaly. After switching the fan speed to one and stabilising its speed, the value of the anomaly score decreased to 1.3, indicating that the fan was working correctly according to the algorithm. For speeds two and three, the average value of the anomaly score was 20 and 10, respectively because these speeds were considered as anomalies.

Before the second experiment, another 16 points were collected to the reservoir for speed two, and the experiment was repeated. Similarly, as before, after stabilising the speeds to one and two, the value of the anomaly score was 1, indicating the device's correct operation. The increase of the value of the anomaly index in the 32nd second of the experiment is due to the transient state of the fan operation during the speed change.

Before the 3rd experiment, another 16 points were collected in the reservoir for speed three. However, during the experiment, for speed three (45th to 60th seconds), the value of the anomaly score momentarily increased to the value of 7. It was caused by fan vibration, which was not manifested during learning phase. To overcome this, more points should be added to the reservoir.

In summary, the presented solution and prototype implementation indicates that it is possible to develop anomaly detection algorithms running on the MCU. Moreover, it is possible to train them directly on the device, adapting to the specific application and device.

\section{Conclusion}
In the paper, we presented the concept of an online anomaly detection algorithm for IoT devices enabling time series analysis from sensors such as an accelerometer. By using reservoir sampling, we reduced the size of the training set by selecting a statistical sample of the time series. It allows the LOF algorithm with quadratic computational complexity to be executed directly on the MCU.

The presented algorithms are implemented as an extension of the FogML tool\footnote{https://github.com/tszydlo/FogML}. The example application is implemented for the Arduino environment\footnote{https://github.com/tszydlo/FogML-Arduino}. It also enables the classification of the operating status of the device. In the case of a fan, this is the speed classification. Additional information is included in the source code.

The presented concept can be extended to other anomaly detection algorithms operating on a set of points such as K-means. The other research area involves the development of strategies and policies for replacing points in the reservoir to follow the concept drift and the ageing of the devices being monitored.


\bibliographystyle{unsrt}  
\bibliography{references}

\end{document}